%% file: rotlstm-thesis.tex
\documentclass{article}

\usepackage[preprint]{nips_2018}

\usepackage[utf8]{inputenc} 
\usepackage[T1]{fontenc}    
\usepackage{hyperref}       
\usepackage{url}            
\usepackage{booktabs}       
\usepackage{amsfonts}       
\usepackage{nicefrac}       
\usepackage{microtype}      
\usepackage{graphicx, wrapfig, subcaption, setspace}
\usepackage[nolist]{acronym}
\usepackage{textcomp}
\usepackage{amsmath}

\graphicspath{{figures/}}   

\title{RotLSTM: rotating memories in Recurrent Neural Networks}

\author{
  Vlad Velici \\
  School of Electronics and Computer Science\\
  University of Southampton\\
  Southampton, UK
  \And
  Adam Pr\"ugel-Bennett \\
  School of Electronics and Computer Science \\
  University of Southampton \\
  Southampton, UK
}

\DeclareMathAlphabet{\mat}{OT1}{cmss}{bx}{n}

\begin{document}

\maketitle

\begin{abstract}
Long Short-Term Memory (LSTM) units have the ability to memorise and use long-term dependencies between inputs to generate predictions on time series data. We introduce the concept of modifying the cell state (memory) of LSTMs using rotation matrices parametrised by a new set of trainable weights. This addition shows significant increases of performance on some of the tasks from the bAbI dataset.
\end{abstract}

\begin{acronym}
  \acro{mAP}[mAP]{mean average precision}
  \acro{cnn}[CNN]{Convolutional Neural Network}
  \acro{rpn}[RPN]{Region Proposal Network}
  \acro{rnn}[RNN]{Recurrent Neural Network}
  \acro{RNN}[RNN]{Recurrent Neural Network}
  \acro{LSTM}[LSTM]{Long Short-Term Memory}
  \acro{NLP}[NLP]{Natural Language Processing}
  \acro{GRU}[GRU]{Gated Recurrent Unit}
\end{acronym}

\section{Introduction}
\label{rot:sec:introduction}

In the recent years, \acp{rnn} have been successfully used to tackle problems with data that can be represented in the shape of time series. Application domains include \ac{NLP} (translation~\citep{DBLP:journals/corr/RoscaB16}, summarisation~\citep{DBLP:journals/corr/NallapatiXZ16}, question answering and more), speech recogition \citep{DBLP:journals/corr/HannunCCCDEPSSCN14, graves2013speech}, text to speech systems \citep{arik2017deep}, computer vision tasks \citep{stewart2016end, wu2017visual}, and differentiable programming language interpreters~\citep{riedel2016programming, rocktaschel2017end}.

An intuitive explanation for the success of RNNs in fields such as natural language understanding is that they allow words at the beginning of a sentence or paragraph to be memorised.  This can be crucial to understanding the semantic content.  Thus in the phrase \emph{"The cat ate the fish"} it is important to memorise the subject (cat).  However, often later words can change the meaning of a senstence in subtle ways.  For example, \emph{"The cat ate the fish, \textbf{didn't it}"} changes a simple statement into a question.  In this paper, we study a mechanism to enhance a standard RNN to enable it to modify its memory, with the hope that this will allow it to capture in the memory cells sequence information using a shorter and more robust representation.

One of the most used \ac{rnn} units is the \ac{LSTM}~\citep{hochreiter1997long}. The core of the LSTM is that each unit has a \emph{cell state} that is modified in a gated fashion at every time step. At a high level, the cell state has the role of providing the neural network with \emph{memory} to hold long-term relationships between inputs. There are many small variations of LSTM units in the literature and most of them yield similar performance~\citep{greff2017lstm}.

The memory (cell state) is expected to encode information necessary to make the next prediction. Currently the ability of the \acp{LSTM} to rotate and swap memory positions is limited to what can be achieved using the available gates. In this work we introduce a new operation on the memory that explicitly enables rotations and swaps of pairwise memory elements. Our preliminary tests show performance improvements on some of the bAbI tasks~\citep{DBLP:journals/corr/WestonBCM15} compared with \ac{LSTM} based architectures.

\section{Related work}

Adding rotations to RNN cells has been previously studied in \cite{DBLP:journals/corr/HenaffSL16}. They do compare their results to LSTM models but do not directly add a new rotation gates to the LSTM like in our work. Their results suggest that rotations do not generalise well for multiple tasks.

Using unitary constraints for weight matrices in RNN units is shown in \cite{DBLP:journals/corr/ArjovskySB15}, where they use the weights in the complex domain (they show a way to only use real numbers in implementation). The EURNN \citep{DBLP:conf/icml/JingSDPSLTS17} presents a more efficient method of parametrising a weights matrix such that it is unitary and uses this parametrisation to add unitary constraints to weights of RNN cells. A similar work is GORU \citep{DBLP:journals/corr/JingGPSTSB17} which adds the same unitary constraints on one of the gate weights in the GRU cell.

A new RNN cell, the Rotational Unit of Memory (RUM) \citep{DBLP:journals/corr/abs-1710-09537}, adds a rotation operation in a gated RNN cell. The rotation parametrisation is passed through to the next cell, similar to how the memory state of an LSTM is. They show increased performance on the bAbI tasks when compared to LSTM and GORU based models.

\section{The rotation gate}
\label{rot:sec:rotation-gate}

In this section we introduce the idea of adding a new set of parameters for the \ac{rnn} cell that enable rotation of the cell state. The following subsection shows how this is implemented in the \ac{LSTM} unit.

One of the key innovations of \acp{LSTM} was the introduction of gated modified states so that if the gate neuron $i$ is saturated then the memory $c_i(t-1)$ would be unaltered. That is, $c_i(t-1) \approx c_i(t)$ with high accuracy. The fact that the amplification factor is very close to 1 prevents the memory vanishing or exploding over many epochs.

To modify the memory, but retain an amplification factor of 1 we take the output after appling the forget and add gates (we call it $\mathbf{d_t}$), and apply a rotation matrix $\mat{U}$ to obtain a modified memory $\mathbf{c}_t=\mat{U}\mathbf{d}_t$. Note that, for a rotation matrix $\mat{U}^T\mat{U} = \mat{I}$ so that $\lVert \mathbf{d_t} \rVert = \lVert \mathbf{c_t} \rVert$.

We parametrise the rotation by a vector of angles
\begin{equation}
	\mathbf{u} = 2\pi\sigma(\mat{W}_{\text{rot}} \mathbf{x} + \mathbf{b}_\text{rot}),
\end{equation}
where $\mat{W}_{\text{rot}}$ is a weight matrix and $\mathbf{b}_\text{rot}$ is a bias vector which we learn along with the other parameters. $\mathbf{x}$ is the vector of our concatenated inputs (in \acp{LSTM} given by concatenating the input for the current timestep with the output from the previous time step).

A full rotation matrix is parametrisable by $n(n-1)/2$ parameters (angles). Using all of these would introduce a huge number of weights, which is likely to over-fit. Instead, we have limited ourselves to considering rotations between pairs of inputs $d_i(t)$ and $d_{i+1}(t)$.

Our rotation matrix is a block-diagonal matrix of 2D rotations \begin{equation}
\label{eq:umat}
	\mat{U}(\mathbf{u}) = \begin{bmatrix}\
		\cos u_1 & -\sin u_1 & &  &  \\
		\sin u_1 & \cos u_1 &  &  &  \\
		 &  & \ddots &  &  \\
		 & &  & \cos u_{n/2} & -\sin u_{n/2} \\
		 &  &  & \sin u_{n/2} & \cos u_{n/2} \
	\end{bmatrix},
\end{equation}
where the cell state is of size $n$. Our choice of rotations only needs $n/2$ angles.

\subsection{RotLSTM}

In this section we show how to add memory rotation to the LSTM unit. The rotation is applied after the forget and add gates and before using the current cell state to produce an output.

\begin{figure}[ht]
\centering
\includegraphics[width=\linewidth]{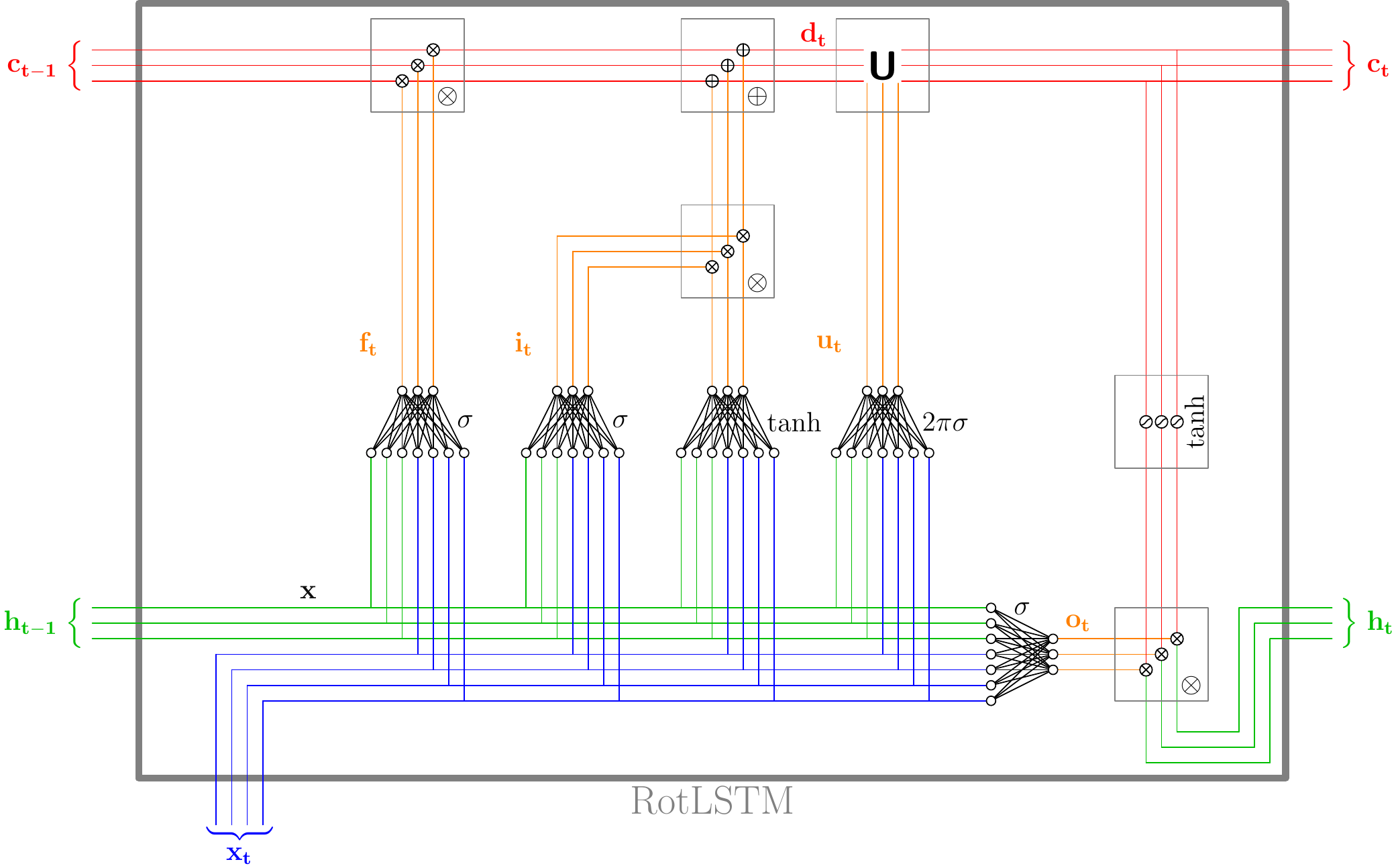}
\caption[RotLSTM unit diagram]{RotLSTM diagram. $\mathbf{x}$ is the concatenation of $\mathbf{h_{t-1}}$ and $\mathbf{x_t}$ in the diagram (green and blue lines). Note that this differs from a regular LSTM by the introduction of the network producing angles $\mathbf{u_t}$ and the rotation module marked $\mat{U}$. In the diagram input size is 4 and cell state size is 3.}
\label{fig:RotLSTM}
\end{figure}

The RotLSTM equations are as follows:
\begin{align}
	\mathbf{x} &= [\mathbf{h_{t-1}},  \mathbf{x_t}], \\
	\mathbf{f_t} &= \sigma(\mat{W}_{\text{f}} \mathbf{x} + \mathbf{b}_\text{f}), \\
	\mathbf{i_t} &= \sigma(\mat{W}_{\text{i}} \mathbf{x} + \mathbf{b}_\text{i}), \\
	\mathbf{o_t} &= \sigma(\mat{W}_{\text{o}} \mathbf{x} + \mathbf{b}_\text{o}), \\
	\mathbf{u_t} &= 2\pi\sigma(\mat{W}_{\text{rot}} \mathbf{x} + \mathbf{b}_\text{rot}), \\
	\mathbf{d_t} &= \mathbf{f_t} \circ \mathbf{c_{t-1}} + \mathbf{i_t} \circ \tanh(\mat{W}_{\text{c}} \mathbf{x} + \mathbf{b}_\text{c}), \\
	\mathbf{c_t} &= \mat{U}(\mathbf{u_t})\mathbf{d_t}, \\
	\mathbf{h_t} &= \mathbf{o_t} \circ \tanh(\mathbf{c_t}),
\end{align}
where $\mat{W_{\{\text{f}, \text{i}, \text{o}, \text{rot}, \text{c}\}}}$ are weight matrices, $\mathbf{b}_{\{\text{f}, \text{i}, \text{o}, \text{rot}, \text{c}\}}$ are biases ($\mat{W}$s and $\mathbf{b}$s learned during training), $\mathbf{h_{t-1}}$ is the previous cell output, $\mathbf{h_t}$ is the output the cell produces for the current timestep, similarly $\mathbf{c_{t-1}}$ and $\mathbf{c_t}$ are the cell states for the previous and current timestep, $\circ$ is element-wise multiplication and $[\cdot, \cdot]$ is concatenation. $\mat{U}$ as defined in Equation~\ref{eq:umat}, parametrised by $\mathbf{u_t}$. Figure~\ref{fig:RotLSTM} shows a RotLSTM unit in detail.

Assuming cell state size $n$, input size $m$, the RotLSTM has $n(n+m)/2$ extra parameters, a 12.5\% increase (ignoring biases). Our expectation is that we can decrease $n$ without harming performance and the rotations will enforce a better representation for the cell state.

\section{Experiments and results}

To empirically evaluate the performance of adding the rotation gate to \acp{LSTM} we use the toy \ac{NLP} dataset bAbI with 1000 samples per task. The bAbI dataset is composed of 20 different tasks of various difficulties, starting from easy questions based on a single supporting fact (for example: \emph{John is in the kitchen. Where is John? A: Kitchen}) and going to more difficult tasks of reasoning about size (example: \emph{The football fits in the suitcase. The box is smaller than the football. Will the box fit in the suitcase? A: yes}) and path finding (example: \emph{The bathroom is south of the office. The bathroom is north of the hallway. How do you go from the hallway to the office?	A: north, north}). A summary of all tasks is available in Table~\ref{table:babi}. We are interested in evaluating the behaviour and performance of rotations on RNN units rather than beating state of the art.

\input{rotlstm/babi_table}

We compare a model based on RotLSTM with the same model based on the traditional LSTM. All models are trained with the same hyperparameters and we do not perform any hyperparameter tuning apart from using the sensible defaults provided in the Keras library and example code~\citep{chollet2015keras}.

For the first experiment we train a LSTM and RotLSTM based model 10 times using a fixed cell state size of 50. In the second experiment we train the same models but vary the cell state size from 6 to 50 to assess whether the rotations help our models achieve good performance with smaller state sizes. We only choose even numbers for the cell state size to make all units go through rotations.

\subsection{The model architecture}

\begin{figure}[ht]
	\begin{center}
		\includegraphics[width=\linewidth]{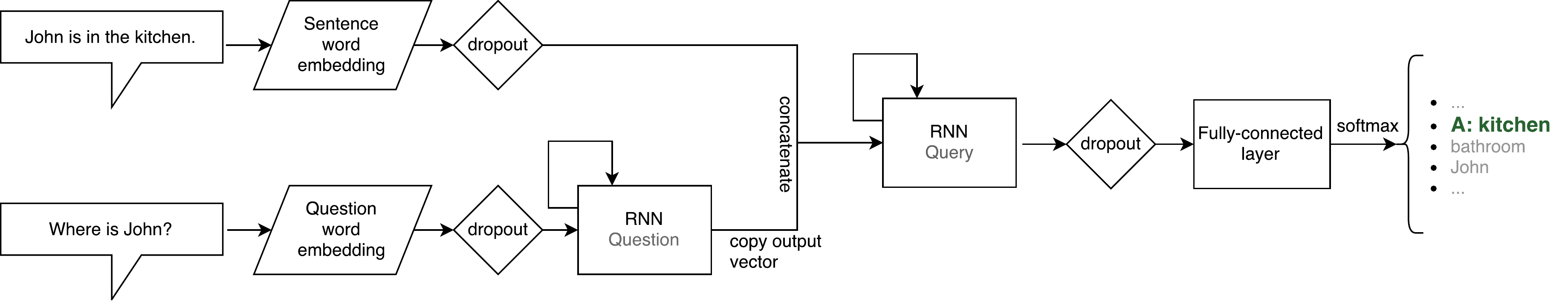}
	\end{center}
	\caption[Model architecture for RotLSTM experiments]{Model architecture for the bAbI dataset. The RNN is either LSTM or RotLSTM.}
	\label{fig:model_arch}
\end{figure}

The model architecture, illustrated in Figure~\ref{fig:model_arch}, is based on the Keras example implementation\footnote{Available at \url{https://goo.gl/9wfzr5}.}. This model architecture, empirically, shows better performance than the \ac{LSTM} baseline published in~\cite{DBLP:journals/corr/WestonBCM15}. The input question and sentences are passed thorugh a word embedding layer (not shared, embeddings are different for questions and sentences). The question is fed into an \ac{rnn} which produces a representation of the question. This representation is concatenated to every word vector from the story, which is then used as input to the second RNN. Intuitively, this helps the second \ac{rnn} (Query) to \emph{focus} on the important words to answer the question. The output of the second \ac{rnn} is passed to a fully connected layer with a softmax activation of the size of the dictionary. The answer is the word with the highest activation.

The categorical cross-entropy loss function was used for training. All dropout layers are dropping 30\% of the nodes. The train-validation dataset split used was 95\%-5\%. The optimizer used was Adam with learning rate 0.001, no decay, $\beta_1=0.9$, $\beta_2=0.999$, $\epsilon = 10^{-8}$. The training set was randomly shuffled before every epoch. All models were trained for 40 epochs. After every epoch the model performance was evaluated on the validation and training sets, and every 10 epochs on the test set. We set the random seeds to the same number for reproducibility and ran the experiments 10 times with 10 different random seeds. The source code is available at \url{https://github.com/vladvelici/swaplstm}.

\subsection{Results}

In this subsection we compare the the performance of models based on the LSTM and RotLSTM units on the bAbI dataset.

\input{rotlstm/results-table}

\begin{figure}[htb]
\begin{center}
\includegraphics[width=\linewidth]{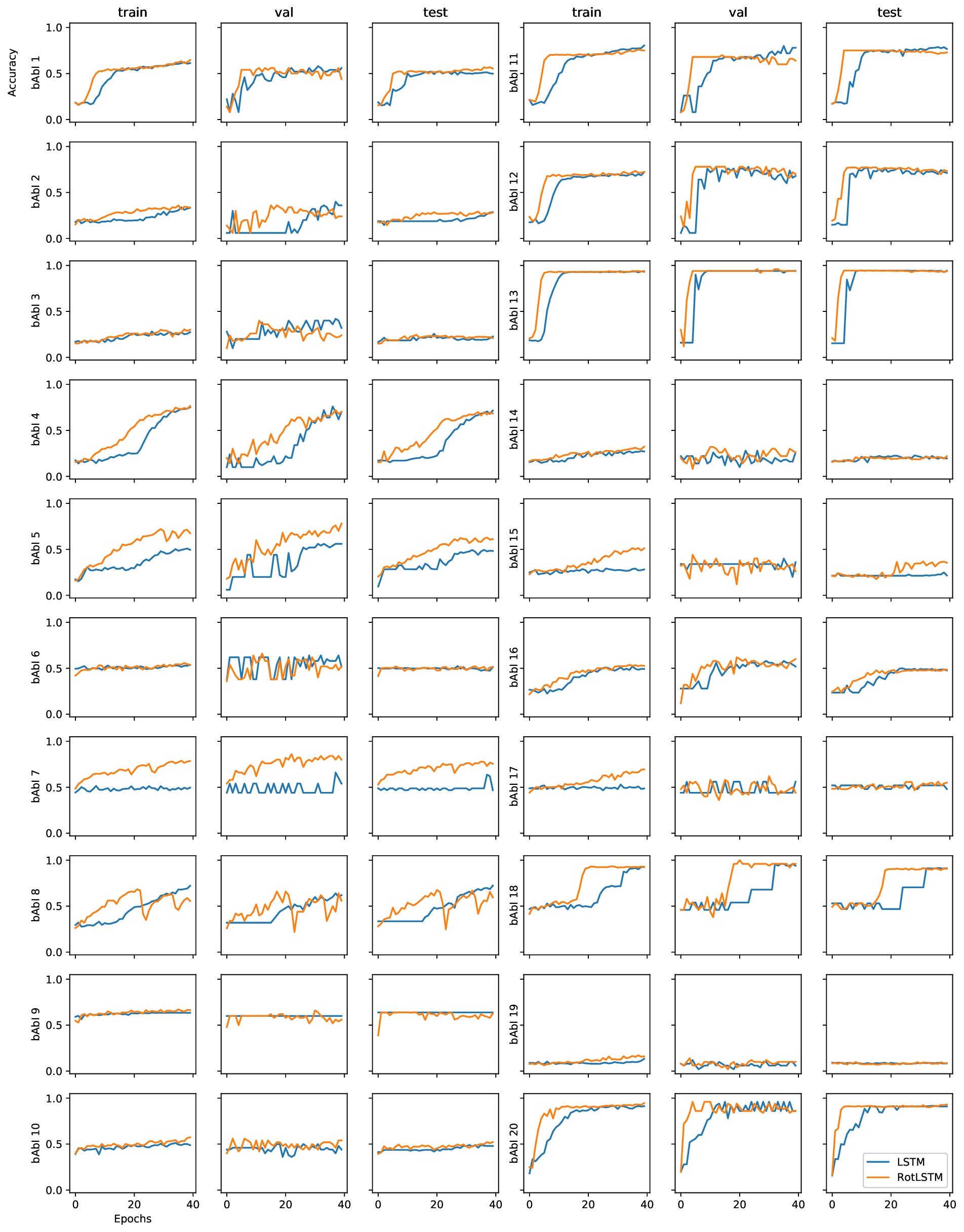}
\end{center}
\caption[LSTM and RotLSTM results on bAbI]{Accuracy comparison on training, validation (val) and test sets over 40 epochs for LSTM and RotLSTM models. The models were trained 10 times and shown is the average accuracy and in faded colour is the standard deviation. Test set accuracy was computed every 10 epochs.}
\label{fig:lstm_converge}
\end{figure}

\begin{figure}[htb]
\begin{center}
\includegraphics[width=\linewidth]{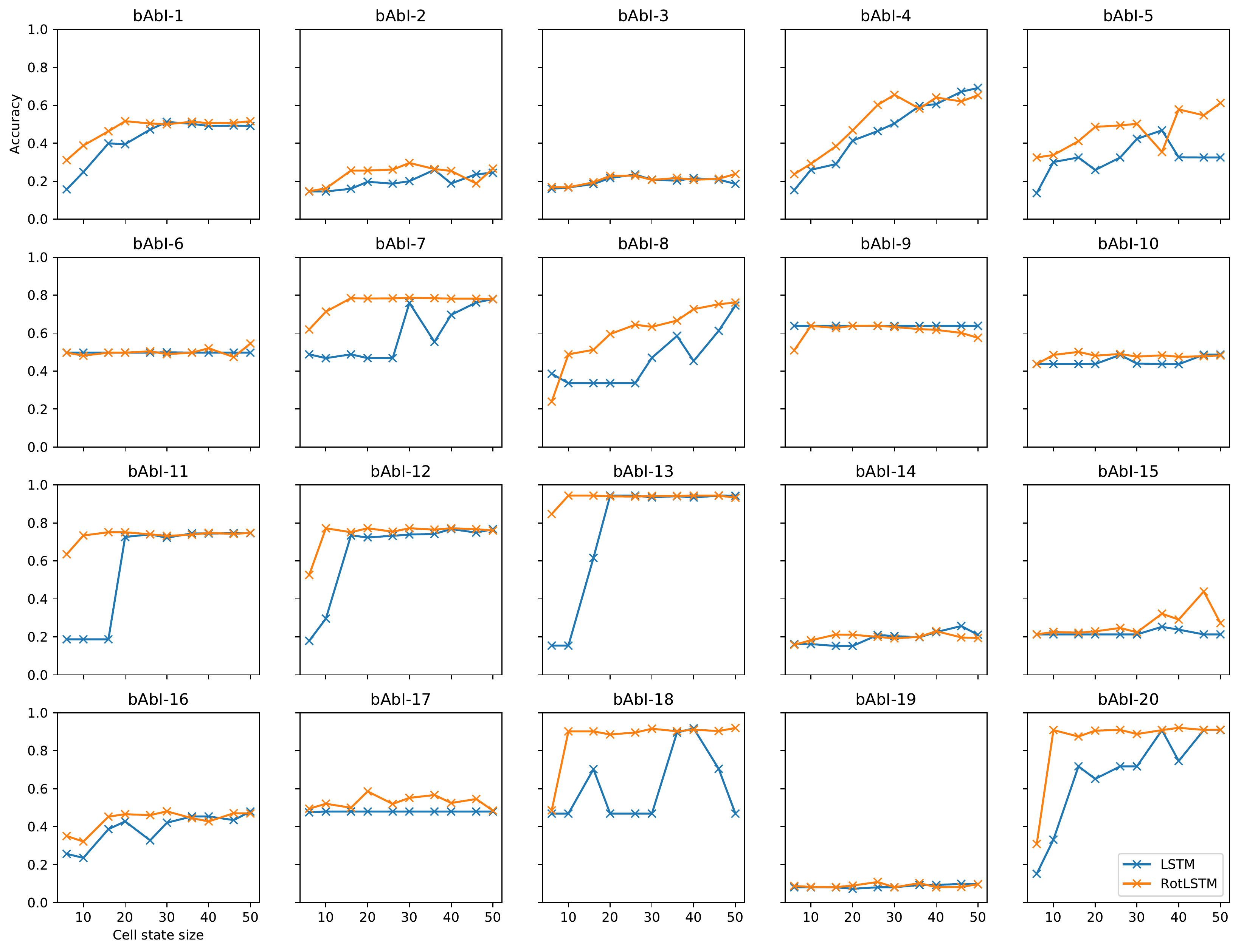}
\end{center}
\caption[LSTM and RotLSTM results with different state cell sizes]{Accuracy on the test set for the LSTM and RotLSTM while varying the cell state size from 6 to 50. The shown numbers are for the epochs with best validation set accuracy.}
\label{fig:state-size-fig-lstm}
\end{figure}

Applying rotations on the unit memory of the LSTM cell gives a slight improvement in performance overall, and significant improvements on specific tasks. Results are shown in Table~\ref{table:results}. The most significant improvements are faster convergence, as shown in Figure~\ref{fig:lstm_converge}, and requiring smaller state sizes, illustrated in Figure~\ref{fig:state-size-fig-lstm}.

On tasks 1 (basic factoid), 11 (basic coreference), 12 (conjunction) and 13 (compound coreference) the RotLSTM model reaches top performance a couple of epochs before the LSTM model consistently. The RotLSTM model also needs a smaller cell state size, reaching top performance at state size 10 to 20 where the LSTM needs 20 to 30. The top performance is, however, similar for both models, with RotLSTM improving the accuracy with up to 2.5\%.

The effect is observed on task 18 (reasoning about size) at a greater magnitude where the RotLSTM reaches top performance before epoch 20, after which it plateaus, while the LSTM model takes 40 epochs to fit the data. The training is more stable for RotLSTM and the final accuracy is improved by 20\%. The RotLSTM reaches top performance using cell state 10 and the LSTM needs size 40. Similar performance increase for the RotLSTM
(22.1\%) is observed in task 5 (three argument relations), reaching top performance around epoch 25 and using a cell state of 50. Task 7 (counting) shows a similar behaviour with an accuracy increase of 14\% for RotLSTM.

Tasks 4 (two argument relations) and 20 (agent motivation) show quicker learning (better performance in the early epochs) for the RotLSTM model but both models reach their top performance after the same amount of traning. On task 20 the RotLSTM performance reaches top accuracy using state size 10 while the LSTM incremetally improves until using state size 40 to 50.

Signs of overfitting for the RotLSTM model can be observed more prominently than for the LSTM model on tasks 15 (basic deduction) and 17 (positional reasoning).

Our models, both LSTM and RotLSTM, perform poorly on tasks 2 and 3 (factoid questions with 2 and 3 supporting facts, respectively) and 14 (time manipulation). These problem classes are solved very well using models that look over the input data many times and use an attention mechanism that allows the model to focus on the relevant input sentences to answer a question \citep{sukhbaatar2015end, kumar2016ask}. Our models only look at the input data once and we do not filter out irrelevant information.

\section{RotGRU}

A popular and successful \ac{RNN} cell is the \ac{GRU} unit \citep{cho2014learning}. Unlike the \ac{LSTM}, the \ac{GRU} unit only has a cell state that can be interpreted as output when needed. Intuitively we want the transformation matrix to enable the unit to learn more compact, richer representations. The rotation is inserted at the \emph{remember} ($\mathbf{r}$) gate to avoid applying a transformation directly on the outputs. The RotGRU equations are the following:
\begin{align}
	\mathbf{x} &= [\mathbf{h_{t-1}},  \mathbf{x_t}], \\
	\mathbf{z_t} &= \sigma(\mat{W}_{\text{z}} \mathbf{x} + \mathbf{b}_\text{z}), \\
	\mathbf{d_t} &= \mathbf{h_{t-1}} \circ \sigma(\mat{W}_{\text{r}} \mathbf{x} + \mathbf{b}_\text{r}), \\
	\mat{U} &= 2\pi\sigma(\mat{W}_{\text{rot}} \mathbf{x} + \mathbf{b}_\text{rot}), \\
	\mathbf{r_t} &= \mat{U}\mathbf{d_t} \\
	\mathbf{\hat{h}_t} &= \tanh(\mat{W}_\text{h}[\mathbf{r_t}, \mathbf{x_t}] + \mathbf{b}_\text{h}), \\
	\mathbf{\mathbf{h}_t} &= (1-\mathbf{z_t}) \circ \mathbf{h_{t-1}} + \mathbf{z_t} \circ \mathbf{\hat{h}_t}.
\end{align}

\begin{figure}[htb]
\begin{center}
\includegraphics[width=\linewidth]{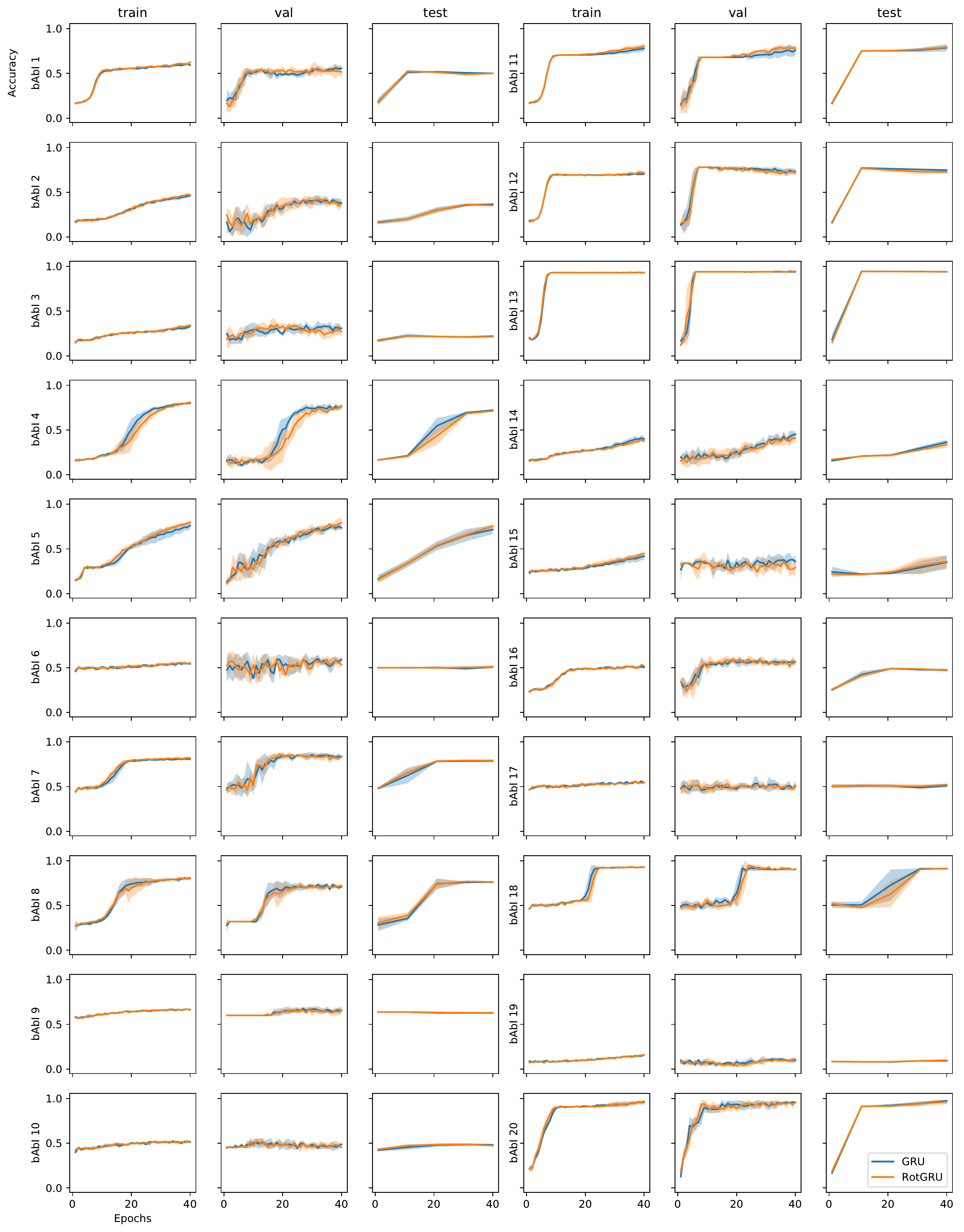}
\end{center}
\caption[GRU and RotGRU results on bAbI]{Accuracy comparison on training, validation (val) and test sets over 40 epochs for GRU and RotGRU models. The models were trained 5 times and shown is the average accuracy and in faded colour is the standard deviation. Test set accuracy was computed every 10 epochs.}
\label{fig:rotgru_converge}
\end{figure}

We show the results of training a RotGRU and a GRU model on the bAbI tasks in Table~\ref{table:rotgru:results} and Figure~\ref{fig:rotgru_converge}. Notice that models based on the GRU cell perform better than the LSTM and RotLSTM cells. From this experiment we also observe that RotGRU performs the same or slightly worse then the initial GRU cell. The RotGRU also requires more computation and introduces more parameters, as a result it takes longer (wall clock) time to train and run for inference.

\input{rotlstm/results-table-gru}

\section{Discussion}

A limitation of the models in our experiments is only applying pairwise 2D rotations. Representations of past input can be larger groups of the cell state vector, thus 2D rotations might not fully exploit the benefits of transformations. Rotating groups of elements and multi-dimensional rotations was not investigated, but could potentially also force the models to learn a more structured representation of the world, similar to how forcing a model to learn specific representations of scenes, as presented in \cite{higgins2017scan}, yields semantic representations of the scene.

In this work we presented preliminary tests for adding rotations to simple models but we only used a toy dataset. The bAbI dataset has certain advantages such as being small thus easy to train many models on a single machine, not having noise as it is generated from a simulation, and having a wide range of tasks of various difficulties. However it is a toy dataset that has a very limited vocabulary and lacks the complexity of real world datasets (noise, inconsistencies, larger vocabularies, more complex language constructs, and so on).

A brief exploration of the angles produced by $\mathbf{u}$ and the weight matrix $\mat{W}_\text{rot}$ show that $\mathbf{u}$ does not saturate, thus rotations are in fact applied to our cell states and do not converge to 0 (or 360 degress). A more in-depth qualitative analysis of the rotation gate might give a better insight into what the rotations are actually achieving. Peeking into the activations of our rotation gates could help understand the behaviour of rotations and to what extent they help (or do not) better represent long-term memory.

A very successful and popular mutation of the \ac{LSTM} is the \ac{GRU} unit \citep{cho2014learning}. The GRU only has an output as opposed to both a cell state and an output and uses fewer gates. We showed that adding similar rotations to GRU cells does not show the same performance improvements, and that the baseline GRU model outperforms both LSTM and RotLSTM models in most cases.

\section{Conclusion}

We have introduced a novel gating mechanism for RNN units that enables applying a parametrised transformation matrix to the cell state. We picked pairwise 2D rotations as the transformation and shown how this can be added to the popular \ac{LSTM} units to create what we call RotLSTM.

We trained a simple model using RotLSTM units and compared them with the same model based on LSTM units. We show that for the LSTM-based architectures adding rotations has a positive impact on most bAbI tasks, making the training require fewer epochs to achieve similar or higher accuracy. On some tasks the RotLSTM model can use a lower dimensional cell state vector and maintain its performance.

Significant accuracy improvements of approximatively 20\% for the RotLSTM model over the LSTM model are visible on bAbI tasks 5 (three argument relations) and 18 (reasoning about size).

The same improvements could not be replicated into the GRU cell when adding our rotation gate, and models based on the GRU cell outperform both the LSTM and RotLSTM on most tasks.

\bibliography{nips_2018}
\bibliographystyle{nips_2018}

\end{document}

%% file: rotlstm/babi_table.tex
\begin{table}[tbp]
\centering
\caption{bAbI dataset tasks.}
\label{table:babi}
\resizebox{\textwidth}{!}{%
\begin{tabular}{@{}rlrl@{}}
\multicolumn{1}{l}{\#} & Description                                  & \multicolumn{1}{l}{\#} & Description                        \\ \midrule
\textbf{1}             & Factoid QA with one supporting fact & \textbf{11}            & Basic coreference                  \\
\textbf{2}             & Factoid QA with two supporting facts         & \textbf{12}            & Conjunction                        \\
\textbf{3}             & Factoid QA with three supporting facts       & \textbf{13}            & Compound coreference               \\
\textbf{4}             & Two argument relations: subject vs. object   & \textbf{14}            & Time manipulation                  \\
\textbf{5}             & Three argument relations                     & \textbf{15}            & Basic deduction                    \\
\textbf{6}             & Yes/No questions                             & \textbf{16}            & Basic induction                    \\
\textbf{7}             & Counting                                     & \textbf{17}            & Positional reasoning               \\
\textbf{8}             & Lists/Sets                                   & \textbf{18}            & Reasoning about size               \\
\textbf{9}             & Simple Negation                              & \textbf{19}            & Path finding                       \\
\textbf{10}            & Indefinite Knowledge                         & \textbf{20}            & Reasoning about agent's motivation \\ \bottomrule
\end{tabular}%
}
\end{table}

%% file: rotlstm/results-table.tex
\begin{table}[tbp]
\centering
\caption[LSTM and RotLSTM results on bAbI]{Performance comparison on the bAbI dataset. Values are \% average accuracy on test set $\pm$standard deviation taken from training each model 10 times. For each of the trained models the test accuracy is taken for the epoch with the best validation accuracy (picked from epoch numbers 1, 11, 21, 31, 40 since we only evaluated on the test set for those).}
\label{table:results}
\resizebox{\textwidth}{!}{%
\begin{tabular}{@{}lllllllllll@{}}
\toprule
\multicolumn{1}{r}{Task:} & \multicolumn{1}{c}{1}  & \multicolumn{1}{c}{2}  & \multicolumn{1}{c}{3}  & \multicolumn{1}{c}{4}  & \multicolumn{1}{c}{5}  & \multicolumn{1}{c}{6}  & \multicolumn{1}{c}{7}  & \multicolumn{1}{c}{8}  & \multicolumn{1}{c}{9}  & \multicolumn{1}{c}{10} \\ \midrule
\textbf{LSTM}             & 49.8 $\pm$3.3          & 26.5 $\pm$3.8          & 21.1 $\pm$1.0          & 63.7 $\pm$9.4          & 33.6 $\pm$6.2          & 49.2 $\pm$0.8          & 62.7 $\pm$15.5         & \textbf{68.8 $\pm$8.3} & \textbf{63.9 $\pm$0.2} & 45.2 $\pm$2.0          \\
\textbf{RotLSTM}          & \textbf{52.3 $\pm$1.1} & \textbf{27.1 $\pm$1.3} & \textbf{22.4 $\pm$1.3} & \textbf{65.8 $\pm$3.6} & \textbf{55.7 $\pm$5.5} & \textbf{50.1 $\pm$1.9} & \textbf{76.7 $\pm$3.0} & 66.1 $\pm$6.2          & 61.5 $\pm$1.9          & \textbf{48.1 $\pm$1.1} \\ \midrule
\multicolumn{1}{r}{Task:} & \multicolumn{1}{c}{11} & \multicolumn{1}{c}{12} & \multicolumn{1}{c}{13} & \multicolumn{1}{c}{14} & \multicolumn{1}{c}{15} & \multicolumn{1}{c}{16} & \multicolumn{1}{c}{17} & \multicolumn{1}{c}{18} & \multicolumn{1}{c}{19} & \multicolumn{1}{c}{20} \\ \midrule
\textbf{LSTM}             & 73.6 $\pm$2.2          & 74.3 $\pm$1.7          & 94.4 $\pm$0.2          & \textbf{20.7 $\pm$3.3} & 21.4 $\pm$0.5          & \textbf{48.0 $\pm$2.2} & 48.0 $\pm$0.0          & 70.3 $\pm$20.8         & 8.5 $\pm$0.6           & 87.7 $\pm$4.2          \\
\textbf{RotLSTM}          & \textbf{73.9 $\pm$1.2} & \textbf{76.5 $\pm$1.1} & \textbf{94.4 $\pm$0.0} & 19.9 $\pm$2.0          & \textbf{28.8 $\pm$8.8} & 46.7 $\pm$2.0          & \textbf{54.2 $\pm$3.1} & \textbf{90.5 $\pm$0.9} & \textbf{8.9 $\pm$1.1}  & \textbf{89.9 $\pm$2.4} \\ \bottomrule
\end{tabular}%
}
\end{table}

%% file: rotlstm/results-table-gru.tex
\begin{table}[]
\centering
\caption[GRU and RotGRU results on bAbI]{Performance comparison on the bAbI dataset. Values are \% average accuracy on test set $\pm$standard deviation taken from training each model 10 times. For each of the trained models the test accuracy is taken for the epoch with the best validation accuracy (picked from epoch numbers 1, 11, 21, 31, 40 since we only evaluated on the test set for those).}
\label{table:rotgru:results}
\resizebox{\textwidth}{!}{%
\begin{tabular}{@{}lllllllllll@{}}
\toprule
\multicolumn{1}{r}{Task:} & \multicolumn{1}{c}{1}  & \multicolumn{1}{c}{2}  & \multicolumn{1}{c}{3}  & \multicolumn{1}{c}{4}  & \multicolumn{1}{c}{5}  & \multicolumn{1}{c}{6}  & \multicolumn{1}{c}{7}  & \multicolumn{1}{c}{8}  & \multicolumn{1}{c}{9}  & \multicolumn{1}{c}{10} \\ \midrule
\textbf{GRU}              & \textbf{51.3 $\pm$0.9} & 34.6 $\pm$3.8          & \textbf{21.5 $\pm$1.4} & 69.8 $\pm$1.9          & 67.1 $\pm$5.7          & 49.9 $\pm$1.0          & \textbf{79.2 $\pm$0.5} & \textbf{76.3 $\pm$1.2} & 62.6 $\pm$0.7          & 48.2 $\pm$1.1          \\
\textbf{RotGRU}           & 50.2 $\pm$1.3          & \textbf{35.7 $\pm$1.9} & 21.3 $\pm$1.1          & \textbf{70.8 $\pm$1.2} & \textbf{73.4 $\pm$2.6} & \textbf{50.1 $\pm$1.1} & 78.4 $\pm$2.6          & 75.9 $\pm$2.0          & \textbf{62.8 $\pm$0.4} & \textbf{48.4 $\pm$1.0} \\ \midrule
\multicolumn{1}{r}{Task:} & \multicolumn{1}{c}{11} & \multicolumn{1}{c}{12} & \multicolumn{1}{c}{13} & \multicolumn{1}{c}{14} & \multicolumn{1}{c}{15} & \multicolumn{1}{c}{16} & \multicolumn{1}{c}{17} & \multicolumn{1}{c}{18} & \multicolumn{1}{c}{19} & \multicolumn{1}{c}{20} \\ \midrule
\textbf{GRU}              & 77.7 $\pm$3.3          & \textbf{76.8 $\pm$0.8} & \textbf{94.4 $\pm$0.0} & \textbf{31.4 $\pm$5.6} & \textbf{27.8 $\pm$6.7} & \textbf{48.0 $\pm$1.3} & \textbf{48.9 $\pm$1.7} & \textbf{91.1 $\pm$0.5} & \textbf{8.9 $\pm$0.8}  & 92.3 $\pm$1.9          \\
\textbf{RotGRU}           & \textbf{79.2 $\pm$2.6} & 76.7 $\pm$0.7          & 94.3 $\pm$0.1          & 29.9 $\pm$4.7          & 23.3 $\pm$4.1          & 47.1 $\pm$3.7          & 48.6 $\pm$1.6          & 90.9 $\pm$0.8          & 8.9 $\pm$1.0           & \textbf{93.6 $\pm$2.9} \\ \bottomrule
\end{tabular}%
}
\end{table}